\documentclass[11pt]{article}

\usepackage[preprint]{acl}

\usepackage{times}
\usepackage{latexsym}

\usepackage{booktabs}
\usepackage{tabularx}
\usepackage{array}
\usepackage[table]{xcolor}
\usepackage{multirow}
\usepackage{listings}
\usepackage{tcolorbox}
\usepackage{lipsum}
\tcbuselibrary{breakable} 

\usepackage[T1]{fontenc}

\usepackage[utf8]{inputenc}

\usepackage{microtype}

\usepackage{inconsolata}

\usepackage{graphicx}

\usepackage{courier}
\usepackage{listings}
\title{Towards Trustworthy Report Generation: A Deep Research Agent with Progressive Confidence Estimation and Calibration}

\author{
  \textbf{Yi Yuan\textsuperscript{1,2}\thanks{Work done during an internship at Shanghai Artificial Intelligence Laboratory.}},
  \textbf{Xuhong Wang\textsuperscript{1}},
  \textbf{Shanzhe Lei\textsuperscript{1}\thanks{Corresponding author}}
  \\
  \\
  \textsuperscript{1}Shanghai Artificial Intelligence Laboratory,
  \textsuperscript{2}Southeast University
  \\
  \small{
  \texttt{yiyuan@seu.edu.cn, \{wangxuhong, leishanzhe\}@pjlab.org.cn}
    }
}

\begin{document}
\maketitle
\begin{abstract}

As agent-based systems continue to evolve, deep research agents are capable of automatically generating research-style reports across diverse domains. While these agents promise to streamline information synthesis and knowledge exploration, existing evaluation frameworks—typically based on subjective dimensions—fail to capture a critical aspect of report quality: \textbf{Trustworthiness}. In open-ended research scenarios where ground-truth answers are unavailable, current evaluation methods cannot effectively measure the epistemic confidence of generated content, making calibration difficult and leaving users susceptible to misleading or hallucinated information. To address this limitation, we propose a novel deep research agent that incorporates progressive confidence estimation and calibration within the report generation pipeline. Our system leverages a Deliberative Search Model, featuring deep retrieval and multi-hop reasoning to ground outputs in verifiable evidence while assigning confidence scores to individual claims. Combined with a carefully designed workflow, this approach produces trustworthy reports with enhanced transparency. Experimental results and case studies demonstrate that our method substantially improves interpretability and significantly increases user trust.

\end{abstract}

\section{Introduction}




\begin{figure}[h]
    \centering
    \includegraphics[width=1.0\columnwidth]{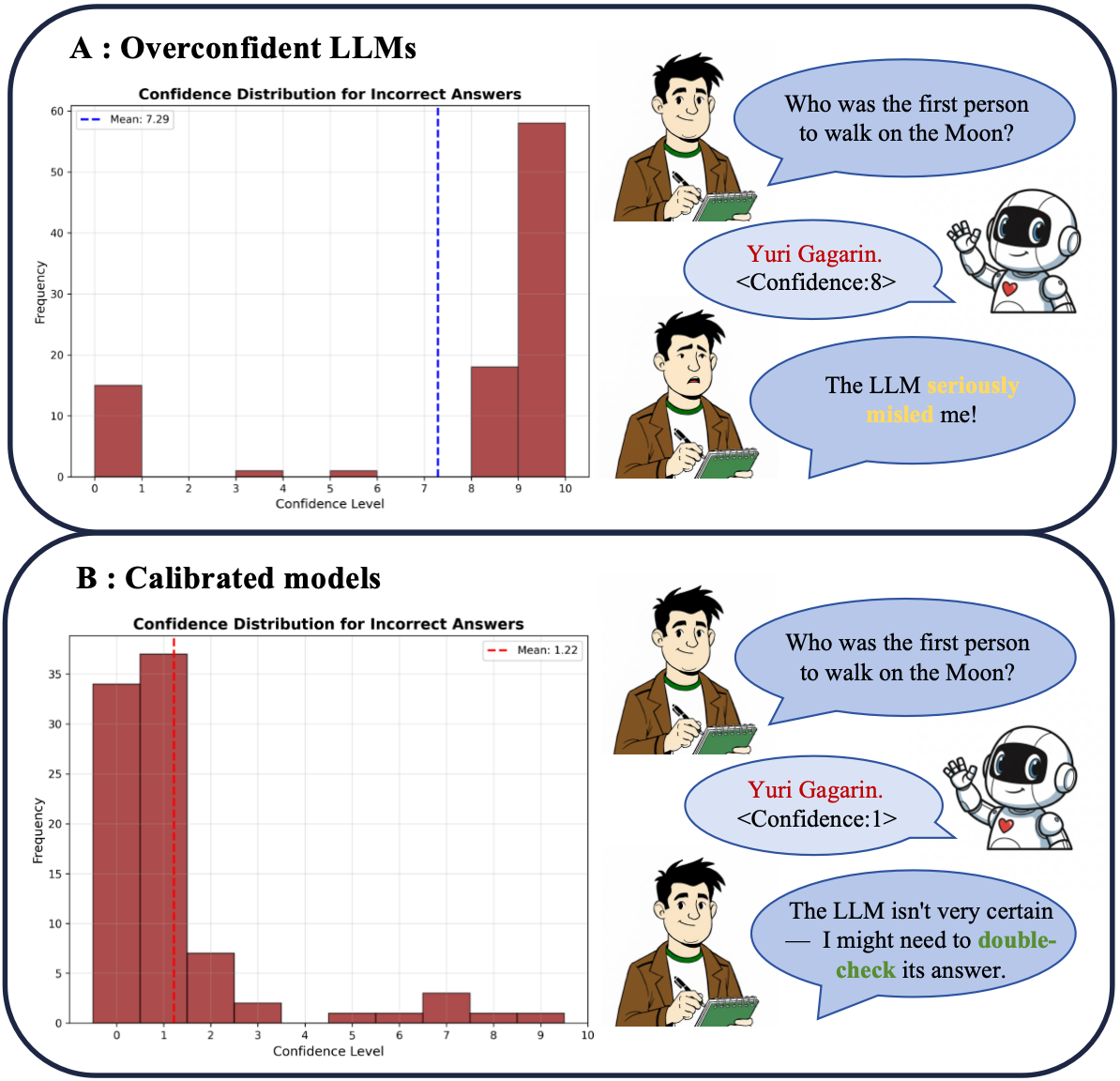}
    \caption{
        Illustration of overconfidence in question answering. When asked \textit{``Who was the first person to walk on the Moon?''}, an overconfident LLM might respond with an incorrect but assertive answer such as \textit{``Yuri Gagarin''} who was the first human in space but never walked on the Moon, which can mislead users. In contrast, a calibrated model may still provide an incorrect answer, but with lower confidence, encouraging users to verify the information.
    }
    \label{fig:overconfidence}
\end{figure}

With the rapid advancement of large language models (LLMs) and agent-based systems, an increasing number of deep research \cite{rodegast2024using, liang2024fourier, yu2023dsformer, li2023camel, qian2023chatdev} have been developed to automatically generate research-style reports across diverse domains. These systems promise to alleviate the burden of information synthesis and knowledge exploration by producing comprehensive, task-specific outputs. However, evaluating the quality and trustworthiness of these generated reports remains a significant challenge. Current evaluation practices typically rely on four subjective dimensions: Comprehensiveness (the breadth and depth of topic coverage), Insight/Depth (the analytical quality and originality of insights), Instruction-Following (adherence to the given task or prompt), and Readability (clarity, organization, and presentation)\cite{du2025deepresearch}. While these metrics provide a general framework for assessing output quality, they fall short in one critical respect: \textbf{they do not offer a reliable measure of trustworthiness}.

This limitation becomes particularly pronounced in open-ended research tasks, where there are no ground-truth answers and constructing reliable evaluation benchmarks is inherently difficult. As a result, LLM-based research agents may generate reports that appear coherent and insightful, yet contain hallucinated information or unsupported claims that are difficult for users to verify. These issues are especially concerning in high-stakes domains such as finance, healthcare, and policy-making, where users often rely on such outputs to support real-world decisions.

As illustrated in Figure~\ref{fig:overconfidence}, the problem of overconfidence is well-documented in the question-answering (QA) domain, where models frequently produce incorrect answers with high certainty. In QA tasks, the presence of ground-truth labels enables the use of calibration techniques\cite{luo2025your, yang2023calibration, manggala2024qa} to mitigate these risks. However, in the context of report generation—particularly for open-domain and long-form outputs—such calibration strategies are largely infeasible due to the absence of definitive reference answers. Consequently, current systems lack mechanisms to evaluate or correct the epistemic confidence of their outputs, leaving users vulnerable to confidently presented but potentially misleading content.

To address this challenge, we argue that modeling of uncertainty and calibration is essential for trustworthy report generation. While direct calibration is difficult due to the absence of ground truth, we propose to incorporate progressive confidence estimation and calibration within the report generation pipeline. Specifically, we decompose report generation into a sequence of QA-style subtasks, each focused on a specific, verifiable query. This allows us to leverage pretrained QA models that are better suited for evidence-grounded generation and confidence estimation. By aligning the strengths of QA calibration with the demands of open-ended report writing, this modular approach introduces finer-grained control over reliability, making it possible to assess and communicate the trustworthiness of individual claims within the report. In doing so, it establishes a practical foundation for uncertainty-aware report generation systems that are both interpretable and robust.


Building on this insight, we introduce a novel deep research agent designed to extend the concept of trustworthiness from QA to full report generation. Our system integrates progressive confidence estimation and calibration mechanisms into the generation pipeline. It leverages a Deliberative Search Model, featuring deep retrieval and multi-hop reasoning to ground its outputs in verifiable sources while assigning confidence scores that reflect the epistemic reliability of individual sections or assertions. In addition, we have carefully designed a three-stage framework for automatic report generation, integrating planning, retrieval, and synthesis. This architecture improves not only the trustworthiness of report generation but also its transparency and adaptability, allowing for more interactive and accountable research workflows.

Through a series of experiments and case studies, we demonstrate that incorporating trustworthiness modeling—via both source-grounded reasoning and uncertainty-aware generation—can substantially improve user confidence in the generated outputs and enhance their practical utility in downstream applications.

\section{Relevant Research}
\subsection{Deep Search \& Deep Research}
Large language models augmented with external tools or knowledge have evolved beyond static retrieval-augmented generation into deep search agents that iteratively query, read, and reason over multiple documents and webs. These agents engage in multi-turn search-read-infer loops, dynamically planning queries and integrating retrieved evidence into chain-of-thought reasoning for complex information needs \cite{xi2025survey, huang2025deep}. Recent systems improve this process through self-refinement and structured memory: for example, new frameworks use reflection and mind-map knowledge graphs to correct errors and maintain coherence across long reasoning chains with web search and other tools \cite{wu2025agentic, guan2025deeprag}.

In parallel, deep research paradigms orchestrate structured multi-step workflows - often via specialized sub-agents or modular planning - to decompose broad tasks and synthesize comprehensive outputs. Multi-agent collaborations can divide labor(e.g., parallel document analyses or cross-checking) and are coordinated by a high-level planner to produce thorough research reports beyond a single-turn chatbot is capacity \cite{zhang2024chain, huang2025manusearch, xu2025comprehensive}. Researchers have introduced transparent open-source agents following this approach \cite{huang2025manusearch} and demontrated that even complex "research" queries benefit from hierarchical planning and tool use. To evaluate these emergent abilities, new benchmarks have been proposed: BrowseComp and BrowseComp-ZH \cite{wei2025browsecomp, zhou2025browsecomp} which chanllenges agents to locate hard-to-find factual information through sustained browsing, and Deep Research Bench provides 100 PhD-level research tasks and two novel evaluation frameworks-RACE and FACT-for assessing report quality and citation accuracy.\cite{mialon2023gaia}.

Overall, early studies underline the promise of these deep search/research frameworks thile also highlighting challenges in aligning retrieval, reasoning, and planing at scale \cite{liang2025reasoning}.

\subsection{Confidence Elicitation in LLMs}

Verbalized Confidence: Lin et al. \cite{lin2022teaching} first taught GPT-3 to output calibrated verbal confidence levels (e.g., "90\% confidence") alongside its answers. Subsequent work, including methods by Yang et al. \cite{yang2024verbalized} and Chen et al. \cite{chen2023quantifying}, demonstrated that with appropriate prompting, LLMs can self-report probabilities that align with correctness. However, these verbalized scores often remain overconfident without further alignment.

Consistency-Based Methods: A different approach infers confidence from answer consistency. Xiong et al. \cite{xiong2023can} show that aggregating multiple outputs from diverse prompts improves calibration. Additionally, methods like self-consistency decoding \cite{taubenfeld2025confidence} and semantic perturbations \cite{lyu2025calibrating} further enhance confidence by measuring agreement between answers or reasoning paths.

External Predictors: Another strategy involves training separate models to estimate the LLM’s correctness. Mielke et al. \cite{mielke2022reducing} used a post-hoc calibrator.

Our framework integrates these approaches to provide a unified, black-box confidence elicitation model for deep research pipelines.


\section{Methodology}
\begin{figure*}[t]
  \centering
  \includegraphics[width=\textwidth]{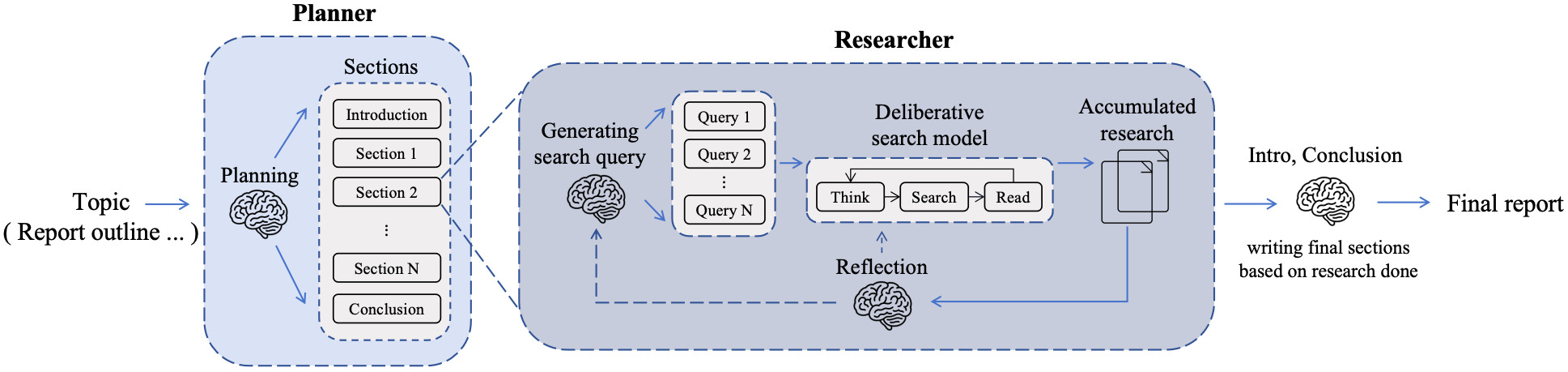}  
  \caption{A three-stage framework for autonomous trustworthy report generation, consisting of a planning module, multiple research workers and synthesis module.}
  \label{fig:workflow}
\end{figure*}

As shown in Figure~\ref{fig:workflow}, we design an autonomous research agent that integrates deliberative reasoning, confidence estimation, and modular workflow orchestration to enable trustworthy, evidence-grounded report generation. Below, we detail the design of the core deliberative model and the orchestration workflow that governs the report generation process.

\subsection{Deliberative Search Model}

Our deep research framework is built upon a deliberative search model, which serves as the core component of the \textbf{Researcher}. The model is designed to tightly integrate step-by-step reasoning with on-demand external knowledge retrieval~\cite{lab2025safework}. Rather than aggregating large amounts of external information upfront, it adopts a reasoning-first paradigm, in which external evidence is actively sought only when the current reasoning state is assessed as insufficient.

In prior work, this deliberative search model was trained using a constrained reinforcement learning framework that jointly optimizes answer accuracy and confidence behavior. Importantly, the confidence signal used in this model is not derived from heuristic rules or post-hoc statistics. Instead, it is produced by a learned scalar prediction head that shares internal representations with the policy network and is optimized end-to-end under the constrained reinforcement learning objective.

From a conceptual perspective, this confidence head can be understood as learning a state-dependent assessment of evidential support. At each step, the model’s internal state encodes both the ongoing reasoning trace and the external information that has been retrieved and read so far. The confidence prediction therefore reflects how consistently and sufficiently the accumulated evidence supports the current intermediate conclusion, rather than attempting to directly estimate the probability that the final answer is correct. In this sense, confidence functions as an internal reliability signal grounded in the model’s reasoning context.

During inference, the model operates in an iterative loop over a fixed action space consisting of \textit{THINK}, \textit{SEARCH}, and \textit{READ}. Each action transitions the model to a new internal reasoning state and is accompanied by an updated confidence estimate. As a result, the confidence signal is process-level and evolves synchronously with the model’s reasoning and information acquisition steps, instead of being limited to the final answer.

\paragraph{THINK Step} 
In the \textit{THINK} step, the model refines its understanding of the problem, produces an intermediate reasoning result or tentative answer, and formulates a query to guide subsequent information retrieval.

\paragraph{SEARCH Step} 
In the \textit{SEARCH} step, the model retrieves potentially relevant external sources based on this query.

\paragraph{READ Step} 
In the \textit{READ} step, the model selectively ingests information from retrieved sources when they are judged to be informative, incorporating new evidence into the ongoing reasoning process.

As deliberation proceeds and additional evidence is accumulated, the model’s confidence may increase when retrieved information consistently supports the emerging conclusion. Conversely, when external evidence remains insufficient or contradictory, the confidence may decrease, signaling increased uncertainty. Crucially, the model is not explicitly optimized to maximize confidence during inference. Instead, confidence naturally emerges as a byproduct of evidence-grounded reasoning shaped by the constrained training objective.

Finally, the entire deliberative inference process is orchestrated through carefully designed prompts that activate the model’s learned decision-making policy, enabling adaptive control over internal reasoning and external verification.

\subsection{Workflow} 

Inspired by the modular workflow \footnote{\url{https://github.com/langchain-ai/open_deep_research}} designed by LangChain-AI, we develop a three-stage framework for automatic report generation, which integrates planning, retrieval, and synthesis into a coherent workflow. As illustrated in Figure~\ref{fig:workflow}, our framework is composed of three modules: \textbf{Planner}, \textbf{Researcher}, and \textbf{Writer}. Each module plays a distinct role in structuring and generating high-quality research-style reports grounded in external knowledge.


\subsubsection{Planner: Topic Decomposition and Section Planning}
Given a user-defined topic or a high-level report outline, the \textbf{Planner} is responsible for decomposing the task into a structured set of sections. This module mimics the behavior of human researchers who begin by outlining the structure of a report before gathering supporting information.

Specifically, the planner generates a sequence of sections---including \textit{Introduction}, \textit{Section 1, Section 2, ..., Section N}, and \textit{Conclusion}---that define the scope and logical flow of the final document. The planning process also determines the granularity of each section, enabling targeted retrieval in later stages.

\subsubsection{Researcher: Deliberative Search and Knowledge Accumulation}

The \textbf{Researcher} module performs iterative, deliberative search to gather relevant evidence and insights for each planned section. For every subsection, the system generates one or more \textit{search queries} tailored to the topic and context. These queries are formulated by a trained query generator that takes into account both the section title and prior context.

The core of the Researcher is a \textit{Deliberative Search Model}, which operates in a \textit{Think} $\rightarrow$ \textit{Search} $\rightarrow$ \textit{Read} cycle. Throughout this process, a \textbf{Reflection} mechanism is employed to evaluate the sufficiency of the accumulated information. If gaps or inconsistencies are detected, new queries are generated and the cycle continues. The outputs of this module are structured content summaries and fact-rich notes for each section.

\subsubsection{Writer: Final Composition and Report Generation}

Once sufficient content has been gathered for all sections, the \textbf{Writer} module composes the final parts of the report---specifically the \textit{Introduction} and \textit{Conclusion}---using the accumulated research. These sections integrate the broader narrative, providing context and summarization respectively.

Our framework emulates the human research-writing process by integrating structured planning, reflective information retrieval, and final composition. By decomposing the task into these stages, we are able to build a system that is both controllable and capable of producing high-quality, research-grounded outputs.

\section{Experiments}
To comprehensively evaluate the effectiveness and trustworthiness of our proposed framework, we design a \textbf{two-tiered} experimental setup that assesses both the core Deliberative Search capability and the overall report generation quality. At the foundational level, we benchmark the Deliberative Search Model on challenging QA datasets to verify its ability to produce accurate, evidence-grounded answers for decomposed sub-queries. Building upon this, we further evaluate the end-to-end performance of our full report generation pipeline on the DeepResearch Bench (DRB). This hierarchical evaluation allows us to analyze how the reliability of individual QA modules contributes to the overall quality, coherence, and trustworthiness of the generated reports.

\subsection{Deliberative Search on QA Benchmark}
To ensure the trustworthiness of the generated research reports, our framework decomposes a complex topic into a set of smaller sub-queries, each of which is handled by the Deliberative Search Model. As the quality of the final report heavily depends on the reliability of the answers produced by this model, it is crucial to assess the factual accuracy and response quality of the Deliberative Search process itself. Therefore, we evaluate our Deliberative Search model on two of the most representative and challenging QA benchmarks to date: GPQA-Diamond and xBench-DeepSearch.

\paragraph{GPQA-Diamond.}  
GPQA-Diamond is a challenging subset of the GPQA benchmark\cite{rein2024gpqa} designed to evaluate advanced reasoning and generalization capabilities in large language models. Unlike simpler QA tasks, GPQA-Diamond includes complex, multi-hop, and compositional questions that require deep understanding across multiple domains. It serves as a rigorous test for probing the limits of factual reasoning and abstraction in AI systems.

\paragraph{xBench-DeepSearch.}  
xBench-DeepSearch\cite{chen2025xbench} is a benchmark designed to evaluate the deep research and information synthesis capabilities of language models. It features complex, open-ended queries that require multi-step reasoning, cross-document evidence retrieval, and coherent report generation. The benchmark emphasizes models' abilities to perform deliberate search, integrate diverse sources, and produce insightful, structured outputs.

\vspace{0.5em}
In evaluating our Deliberative Search Model, we adopt different evaluation metrics tailored to the distinct objectives of each QA benchmark. For \textbf{GPQA-Diamond}, we use \emph{accuracy} as the primary metric, as the benchmark consists of well-defined multiple-choice questions with objectively correct answers. Accuracy effectively reflects the model's ability to perform precise factual reasoning and select the correct answer from a set of alternatives, making it a suitable indicator of raw reasoning capability.

In contrast, \textbf{xBench-DeepSearch} focuses on open-ended, research-style queries that require multi-step reasoning, synthesis of information across sources, and structured answer generation. In such settings, correctness is often not binary and ground-truth answers are not always definitive. Therefore, we adopt \emph{Expected Calibration Error (ECE)} to evaluate the reliability of the model’s confidence scores. ECE captures how well the model’s predicted probabilities align with the actual correctness of its outputs, providing insights into whether the model is \emph{overconfident} or \emph{underconfident} in its deliberative search process. This is crucial in research-style applications where users rely on the model’s confidence to assess the trustworthiness of the synthesized content.


\paragraph{Results.}  
As shown in Table~\ref{tab:gpqa_results} and Table~\ref{tab:xbench_results}, our model achieves strong performance on both QA benchmarks, demonstrating its effectiveness in handling complex queries through deliberate search. On \textbf{GPQA-Diamond}, the model attains a high accuracy, indicating robust factual reasoning and the ability to resolve challenging multi-hop and compositional questions. On \textbf{xBench-DeepSearch}, the model exhibits a notably low Expected Calibration Error (ECE), reflecting not only high-quality answer synthesis but also well-calibrated confidence estimates—an essential property for supporting downstream decision-making and information aggregation. These results validate the model's capability to retrieve, reason, and integrate information across diverse contexts, establishing a solid foundation for high-quality report generation in more open-ended, research-driven scenarios.

\begin{table}[ht]
    \small
    \centering
    \begin{tabular}{c|c}
        \toprule[1pt]\midrule
        \textbf{Methods} & \textbf{Accuracy} $\uparrow$ \\
        \midrule
        Qwen-VL2.5-72B & 48.99 \\
        GPT-4o & 33.48 \\
        Gemini-2.0-Flash & 32.39 \\
        Gemini-2.0-Flash-Thinking & 40.91 \\
        Gemini-2.5-Pro-Exp & 52.02 \\
        Claude-3-7-Sonnet & 54.04 \\
        Claude-3-7-Sonnet-Thinking & 54.04 \\
        \rowcolor[gray]{0.9}
        \textbf{Ours (Deliberative Search)} & \textbf{61.62} \\
        \midrule\bottomrule[1pt]
    \end{tabular}
    \caption{Accuracy results of various large language models evaluated on the GPQA-Diamond benchmark.}
    \label{tab:gpqa_results}
\end{table}

\begin{table}[ht]
    \small
    \centering
    \begin{tabular}{c|c}
        \toprule[1pt]\midrule
        \textbf{Methods} & $\mathbf{N}_{\mathrm{ECE}} \downarrow$ \\
        \midrule
        Qwen-VL2.5-72B                 & 0.53 \\
        Intern-VL3-78B                  & 0.50 \\
        DeepSeek-R1-Distill-Llama-70B & 0.42 \\
        GPT-4.1                       & 0.47 \\
        GPT-4o                        & 0.39 \\
        Claude-4-Sonnet              & 0.36 \\
        \rowcolor[gray]{.9}
        \textbf{Ours (Deliberative Search)} & \textbf{0.34} \\
        \midrule\bottomrule[1pt]
    \end{tabular}
    \caption{Calibration performance measured by the normalized Expected Calibration Error ($\mathbf{N}_{\mathrm{ECE}}$) on the xBench-DeepSearch benchmark across various large language models.}
    \label{tab:xbench_results}
\end{table}

\subsection{Report Quality on DeepResearch Bench}
To evaluate the effectiveness of our report generation framework, we conduct experiments on the DeepResearch Bench (DRB)\cite{du2025deepresearch}, a recently proposed benchmark designed for assessing the quality of long-form research-style outputs. The dataset consists of 100 PhD-level research tasks, each meticulously crafted by domain experts across 22 distinct fields such as finance, healthcare, technology, and public policy. Each question is accompanied by a language tag (e.g., \texttt{zh}, \texttt{en}) and requires systems to produce informative, structured, and trustworthy long-form responses.

\paragraph{Experimental Setup.}
Our system adopts a multi-stage workflow, for all stages of the pipeline—including \textbf{planning}, \textbf{writing}, and \textbf{reflection}—we employ \texttt{gpt-4o} as the backbone model to ensure consistency and high-quality generation across the pipeline. Specifically:

\begin{itemize}
    \item \textbf{Planning:} Given a user query, the system first generates a high-level outline that identifies major research points, subtopics, and supporting facts to be investigated.
    \item \textbf{Writing:} Each planned subtopic is then expanded into a detailed paragraph or section, guided by the planning phase and executed by the same GPT-4o model.
    \item \textbf{Reflection:} Finally, the system conducts a self-reflection phase to verify factual consistency, coherence, and instruction alignment, optionally revising parts of the draft where necessary.
\end{itemize}

\paragraph{Evaluation.}
To systematically evaluate the quality of generated research reports, we adopt the RACE (Reference-based Adaptive Criteria-driven Evaluation) framework from DeepResearch Bench\cite{du2025deepresearch}. RACE provides a comprehensive multi-dimensional assessment by dynamically generating task-specific evaluation criteria across four key aspects: \textit{Comprehensiveness}, \textit{Insight/Depth}, \textit{Instruction-Following}, and \textit{Readability}. Each generated report is compared against high-quality reference reports through reference-based scoring, enabling a fine-grained and discriminative evaluation. Furthermore, RACE introduces adaptive weighting tailored to each task’s objectives, ensuring that the final scores reflect both the general and task-specific expectations of high-quality research outputs.

\paragraph{Results.}

\begin{table*}[t]
    \begin{center}
        \resizebox{1.9\columnwidth}{!}{
            \begin{tabular}{c|ccccc}
                \toprule[1pt]\midrule
                \textbf{Method} & \textbf{Overall} & \textbf{Comprehensiveness} & \textbf{Insightfulness} & \textbf{Instruction} & \textbf{Readability} \\
                \midrule
                \multicolumn{6}{c}{\textbf{Deep Research Agents}} \\
                \midrule
                Gemini-2.5-Pro Deep Research & \textbf{48.92} & \textbf{48.45} & \textbf{48.30} & 49.29 & \textbf{49.77} \\
                OpenAI Deep Research & 46.45 & 46.46 & 43.73 & \textbf{49.39} & 47.22 \\
                Claude-Researcher & 45.00 & 45.34 & 42.79 & 47.58 & 44.66 \\
                Kimi-Researcher & 44.64 & 44.96 & 41.97 & 47.14 & 45.59 \\
                Doubao-DeepResearch & 44.34 & 44.84 & 40.56 & 47.95 & 44.69 \\
                Perplexity-Research & 40.46 & 39.10 & 35.65 & 46.11 & 43.08 \\
                Grok Deeper Search & 38.22 & 36.08 & 30.89 & 46.59 & 42.17 \\
                \midrule
                \multicolumn{6}{c}{\textbf{LLM with Search Tools}} \\
                \midrule
                Perplexity-Sonar-Reasoning-Pro & 37.76 & 34.96 & 31.65 & 44.93 & 42.42 \\
                Perplexity-Sonar-Reasoning & 37.75 & 34.73 & 32.59 & 44.42 & 42.39 \\
                Claude-3.7-Sonnet w/Search & 36.63 & 35.95 & 31.29 & 44.05 & 36.07 \\
                Perplexity-Sonar-Pro & 36.19 & 33.92 & 29.69 & 43.39 & 41.07 \\
                \rowcolor[gray]{0.9} \textbf{Ours (Report)} & 34.13 & 32.15 & 28.07 & 41.25 & 38.19 \\
                Gemini-2.5-Pro-Preview & 31.90 & 31.75 & 24.61 & 40.24 & 32.76 \\
                GPT-4o-Search-Preview & 30.74 & 27.81 & 20.44 & 41.01 & 37.60 \\
                Perplexity-Sonar & 30.64 & 27.14 & 21.62 & 40.70 & 37.46 \\
                GPT-4.1 w/Search & 29.31 & 25.59 & 18.42 & 40.63 & 36.49 \\
                Gemini-2.5-Flash-Preview & 29.19 & 28.97 & 21.62 & 37.80 & 29.97 \\
                GPT-4o-Mini-Search-Preview & 27.62 & 24.24 & 16.62 & 38.59 & 35.27 \\
                GPT-4.1-Mini w/Search & 26.62 & 22.86 & 15.39 & 38.18 & 34.49 \\
                Claude-3.5-Sonnet w/Search & 23.95 & 21.28 & 16.20 & 32.41 & 29.87 \\
                \midrule\bottomrule[1pt]
            \end{tabular}
        }
    \end{center}
    \vspace{-0.2cm}
    \caption{Comparative results of different deep research agents and LLMs with search tools on the Deep Research Bench across four evaluation dimensions.}
    \label{tab:drb-results}
\end{table*}

 As shown in Table~\ref{tab:drb-results}, compared with several proprietary deep research agents and closed LLMs integrated with multi-step search frameworks, our model achieves \textbf{competitive performance}, ranking around the \textbf{mid-range} across key quality dimensions including \textit{comprehensiveness}, \textit{insightfulness}, and \textit{instruction-following}. Notably, our approach strikes a favorable balance between report quality and reliability, offering outputs that are both coherent and evidence-grounded.

In Appendix~\ref{sec:agent-demonstration}, we showcase our report generation process on a randomly selected topic from the DeepResearch benchmark. Appendix~\ref{sec:report-Comparison} presents case studies comparing our system’s reports with those from baseline models. We further analyze how confidence is expressed throughout the generated reports. Our system consistently assigns higher epistemic confidence to claims supported by well-defined, objective evidence such as quantitative data or scientific facts, while exhibiting lower confidence when addressing more abstract or speculative topics, including philosophical or policy-related discussions. This behavior reflects the model’s discriminative ability to calibrate confidence in accordance with the inherent epistemic uncertainty of the task domain, thereby enhancing the reliability and trustworthiness of the final research outputs.

\section{Conclusion}
In this work, we have presented a novel deep research framework that addresses a fundamental limitation in current report generation systems: the lack of trustworthiness estimation in open-ended, long-form outputs. By decomposing the report generation process into a sequence of verifiable QA-style subtasks, and integrating progressive confidence estimation and calibration mechanisms, our framework bridges the gap between epistemic uncertainty modeling and large-scale, automated research synthesis. Through the use of a deliberative search model and modular planning-synthesis stages, our system provides not only more reliable outputs, but also a clearer representation of the confidence associated with each generated claim. We hope this work lays the foundation for building more transparent, trustworthy, and accountable research agents, and serves as a step toward closing the gap between LLM capabilities and the rigorous demands of human-centered knowledge work.

\bibliography{latex/ref}

\begin{thebibliography}{30}
\providecommand{\natexlab}[1]{#1}

\bibitem[{Chen and Mueller(2023)}]{chen2023quantifying}
Jiuhai Chen and Jonas Mueller. 2023.
\newblock Quantifying uncertainty in answers from any language model and enhancing their trustworthiness.
\newblock \emph{arXiv preprint arXiv:2308.16175}.

\bibitem[{Chen et~al.(2025)Chen, Ren, Liu, Hu, Tian, Xie, Liu, Zhang, Liu, Gong et~al.}]{chen2025xbench}
Kaiyuan Chen, Yixin Ren, Yang Liu, Xiaobo Hu, Haotong Tian, Tianbao Xie, Fangfu Liu, Haoye Zhang, Hongzhang Liu, Yuan Gong, and 1 others. 2025.
\newblock xbench: Tracking agents productivity scaling with profession-aligned real-world evaluations.
\newblock \emph{arXiv preprint arXiv:2506.13651}.

\bibitem[{Du et~al.(2025)Du, Xu, Zhu, Wang, and Mao}]{du2025deepresearch}
Mingxuan Du, Benfeng Xu, Chiwei Zhu, Xiaorui Wang, and Zhendong Mao. 2025.
\newblock Deepresearch bench: A comprehensive benchmark for deep research agents.
\newblock \emph{arXiv preprint}.

\bibitem[{Guan et~al.(2025)Guan, Zeng, Meng, Xin, Lu, Lin, Han, Sun, and Zhou}]{guan2025deeprag}
Xinyan Guan, Jiali Zeng, Fandong Meng, Chunlei Xin, Yaojie Lu, Hongyu Lin, Xianpei Han, Le~Sun, and Jie Zhou. 2025.
\newblock Deeprag: Thinking to retrieve step by step for large language models.
\newblock \emph{arXiv preprint arXiv:2502.01142}.

\bibitem[{Huang et~al.(2025{\natexlab{a}})Huang, Liu, Jiang, Zhang, Yan, Li, and Zhao}]{huang2025manusearch}
Lisheng Huang, Yichen Liu, Jinhao Jiang, Rongxiang Zhang, Jiahao Yan, Junyi Li, and Wayne~Xin Zhao. 2025{\natexlab{a}}.
\newblock Manusearch: Democratizing deep search in large language models with a transparent and open multi-agent framework.
\newblock \emph{arXiv preprint arXiv:2505.18105}.

\bibitem[{Huang et~al.(2025{\natexlab{b}})Huang, Chen, Zhang, Li, Fang, Yang, Li, Shang, Xu, Hao et~al.}]{huang2025deep}
Yuxuan Huang, Yihang Chen, Haozheng Zhang, Kang Li, Meng Fang, Linyi Yang, Xiaoguang Li, Lifeng Shang, Songcen Xu, Jianye Hao, and 1 others. 2025{\natexlab{b}}.
\newblock Deep research agents: A systematic examination and roadmap.
\newblock \emph{arXiv preprint arXiv:2506.18096}.

\bibitem[{Lab et~al.(2025)Lab, Bao, Chen, Chen, Chen, Chen, Chen, Chen, Chen, Cheng et~al.}]{lab2025safework}
Shanghai~AI Lab, Yicheng Bao, Guanxu Chen, Mingkang Chen, Yunhao Chen, Chiyu Chen, Lingjie Chen, Sirui Chen, Xinquan Chen, Jie Cheng, and 1 others. 2025.
\newblock {SafeWork-R1}: Coevolving safety and intelligence under the {AI-45°} law.
\newblock \emph{arXiv preprint arXiv:2507.18576}.

\bibitem[{Li et~al.(2023)Li, Hammoud, Itani, Khizbullin, and Ghanem}]{li2023camel}
Guohao Li, Hasan Hammoud, Hani Itani, Dmitrii Khizbullin, and Bernard Ghanem. 2023.
\newblock Camel: Communicative agents for" mind" exploration of large language model society.
\newblock \emph{Advances in Neural Information Processing Systems}, 36:51991--52008.

\bibitem[{Liang et~al.(2025)Liang, Su, Lin, Wu, Zhao, and Li}]{liang2025reasoning}
Jintao Liang, Gang Su, Huifeng Lin, You Wu, Rui Zhao, and Ziyue Li. 2025.
\newblock Reasoning rag via system 1 or system 2: A survey on reasoning agentic retrieval-augmented generation for industry challenges.
\newblock \emph{arXiv preprint arXiv:2506.10408}.

\bibitem[{Liang et~al.(2024)Liang, Cao, Deng, Dou, and Deng}]{liang2024fourier}
YuJie Liang, Zihan Cao, Shangqi Deng, Hong-Xia Dou, and Liang-Jian Deng. 2024.
\newblock Fourier-enhanced implicit neural fusion network for multispectral and hyperspectral image fusion.
\newblock \emph{Advances in Neural Information Processing Systems}, 37:63441--63465.

\bibitem[{Lin et~al.(2022)Lin, Hilton, and Evans}]{lin2022teaching}
Stephanie Lin, Jacob Hilton, and Owain Evans. 2022.
\newblock Teaching models to express their uncertainty in words.
\newblock \emph{arXiv preprint arXiv:2205.14334}.

\bibitem[{Luo et~al.(2025)Luo, Wang, Li, and Wei}]{luo2025your}
Beier Luo, Shuoyuan Wang, Yixuan Li, and Hongxin Wei. 2025.
\newblock Your pre-trained llm is secretly an unsupervised confidence calibrator.
\newblock \emph{arXiv preprint arXiv:2505.16690}.

\bibitem[{Lyu et~al.(2025)Lyu, Shridhar, Malaviya, Zhang, Elazar, Tandon, Apidianaki, Sachan, and Callison-Burch}]{lyu2025calibrating}
Qing Lyu, Kumar Shridhar, Chaitanya Malaviya, Li~Zhang, Yanai Elazar, Niket Tandon, Marianna Apidianaki, Mrinmaya Sachan, and Chris Callison-Burch. 2025.
\newblock Calibrating large language models with sample consistency.
\newblock In \emph{Proceedings of the AAAI Conference on Artificial Intelligence}, volume~39, pages 19260--19268.

\bibitem[{Manggala et~al.(2024)Manggala, Mastakouri, Kirschbaum, Kasiviswanathan, and Ramdas}]{manggala2024qa}
Putra Manggala, Atalanti Mastakouri, Elke Kirschbaum, Shiva~Prasad Kasiviswanathan, and Aaditya Ramdas. 2024.
\newblock Qa-calibration of language model confidence scores.
\newblock \emph{arXiv preprint arXiv:2410.06615}.

\bibitem[{Mialon et~al.(2023)Mialon, Fourrier, Wolf, LeCun, and Scialom}]{mialon2023gaia}
Gr{\'e}goire Mialon, Cl{\'e}mentine Fourrier, Thomas Wolf, Yann LeCun, and Thomas Scialom. 2023.
\newblock Gaia: a benchmark for general ai assistants.
\newblock In \emph{The Twelfth International Conference on Learning Representations}.

\bibitem[{Mielke et~al.(2022)Mielke, Szlam, Dinan, and Boureau}]{mielke2022reducing}
Sabrina~J Mielke, Arthur Szlam, Emily Dinan, and Y-Lan Boureau. 2022.
\newblock Reducing conversational agents’ overconfidence through linguistic calibration.
\newblock \emph{Transactions of the Association for Computational Linguistics}, 10:857--872.

\bibitem[{Qian et~al.(2023)Qian, Liu, Liu, Chen, Dang, Li, Yang, Chen, Su, Cong et~al.}]{qian2023chatdev}
Chen Qian, Wei Liu, Hongzhang Liu, Nuo Chen, Yufan Dang, Jiahao Li, Cheng Yang, Weize Chen, Yusheng Su, Xin Cong, and 1 others. 2023.
\newblock Chatdev: Communicative agents for software development.
\newblock \emph{arXiv preprint arXiv:2307.07924}.

\bibitem[{Rein et~al.(2024)Rein, Hou, Stickland, Petty, Pang, Dirani, Michael, and Bowman}]{rein2024gpqa}
David Rein, Betty~Li Hou, Asa~Cooper Stickland, Jackson Petty, Richard~Yuanzhe Pang, Julien Dirani, Julian Michael, and Samuel~R. Bowman. 2024.
\newblock \href {https://openreview.net/forum?id=Ti67584b98} {{GPQA}: A graduate-level google-proof q\&a benchmark}.
\newblock In \emph{First Conference on Language Modeling}.

\bibitem[{Rodegast et~al.(2024)Rodegast, Maier, Kneifl, and Fehr}]{rodegast2024using}
Philipp Rodegast, Steffen Maier, Jonas Kneifl, and J{"o}rg Fehr. 2024.
\newblock On using machine learning algorithms for motorcycle collision detection.
\newblock \emph{Discover Applied Sciences}, 6(6):326.

\bibitem[{Taubenfeld et~al.(2025)Taubenfeld, Sheffer, Ofek, Feder, Goldstein, Gekhman, and Yona}]{taubenfeld2025confidence}
Amir Taubenfeld, Tom Sheffer, Eran Ofek, Amir Feder, Ariel Goldstein, Zorik Gekhman, and Gal Yona. 2025.
\newblock Confidence improves self-consistency in llms.
\newblock \emph{arXiv preprint arXiv:2502.06233}.

\bibitem[{Wei et~al.(2025)Wei, Sun, Papay, McKinney, Han, Fulford, Chung, Passos, Fedus, and Glaese}]{wei2025browsecomp}
Jason Wei, Zhiqing Sun, Spencer Papay, Scott McKinney, Jeffrey Han, Isa Fulford, Hyung~Won Chung, Alex~Tachard Passos, William Fedus, and Amelia Glaese. 2025.
\newblock Browsecomp: A simple yet challenging benchmark for browsing agents.
\newblock \emph{arXiv preprint arXiv:2504.12516}.

\bibitem[{Wu et~al.(2025)Wu, Zhu, Liu, Xu, and Jin}]{wu2025agentic}
Junde Wu, Jiayuan Zhu, Yuyuan Liu, Min Xu, and Yueming Jin. 2025.
\newblock Agentic reasoning: A streamlined framework for enhancing llm reasoning with agentic tools.
\newblock In \emph{Proceedings of the 63rd Annual Meeting of the Association for Computational Linguistics (Volume 1: Long Papers)}, pages 28489--28503.

\bibitem[{Xi et~al.(2025)Xi, Lin, Xiao, Zhou, Shan, Gao, Zhu, Liu, Yu, and Zhang}]{xi2025survey}
Yunjia Xi, Jianghao Lin, Yongzhao Xiao, Zheli Zhou, Rong Shan, Te~Gao, Jiachen Zhu, Weiwen Liu, Yong Yu, and Weinan Zhang. 2025.
\newblock A survey of llm-based deep search agents: Paradigm, optimization, evaluation, and challenges.
\newblock \emph{arXiv preprint arXiv:2508.05668}.

\bibitem[{Xiong et~al.(2023)Xiong, Hu, Lu, Li, Fu, He, and Hooi}]{xiong2023can}
Miao Xiong, Zhiyuan Hu, Xinyang Lu, Yifei Li, Jie Fu, Junxian He, and Bryan Hooi. 2023.
\newblock Can llms express their uncertainty? an empirical evaluation of confidence elicitation in llms.
\newblock \emph{arXiv preprint arXiv:2306.13063}.

\bibitem[{Xu and Peng(2025)}]{xu2025comprehensive}
Renjun Xu and Jingwen Peng. 2025.
\newblock A comprehensive survey of deep research: Systems, methodologies, and applications.
\newblock \emph{arXiv preprint arXiv:2506.12594}.

\bibitem[{Yang et~al.(2024)Yang, Tsai, and Yamada}]{yang2024verbalized}
Daniel Yang, Yao-Hung~Hubert Tsai, and Makoto Yamada. 2024.
\newblock On verbalized confidence scores for llms.
\newblock \emph{arXiv preprint arXiv:2412.14737}.

\bibitem[{Yang et~al.(2023)Yang, Dan, Roth, and Lee}]{yang2023calibration}
Yahan Yang, Soham Dan, Dan Roth, and Insup Lee. 2023.
\newblock On the calibration of multilingual question answering llms.
\newblock \emph{arXiv preprint arXiv:2311.08669}.

\bibitem[{Yu et~al.(2023)Yu, Wang, Shao, Sun, Wu, and Xu}]{yu2023dsformer}
Chengqing Yu, Fei Wang, Zezhi Shao, Tao Sun, Lin Wu, and Yongjun Xu. 2023.
\newblock Dsformer: A double sampling transformer for multivariate time series long-term prediction.
\newblock In \emph{Proceedings of the 32nd ACM international conference on information and knowledge management}, pages 3062--3072.

\bibitem[{Zhang et~al.(2024)Zhang, Sun, Chen, Pfister, Zhang, and Arik}]{zhang2024chain}
Yusen Zhang, Ruoxi Sun, Yanfei Chen, Tomas Pfister, Rui Zhang, and Sercan Arik. 2024.
\newblock Chain of agents: Large language models collaborating on long-context tasks.
\newblock \emph{Advances in Neural Information Processing Systems}, 37:132208--132237.

\bibitem[{Zhou et~al.(2025)Zhou, Leon, Ying, Zhang, Shao, Ye, Chong, Jin, Xie, Cao et~al.}]{zhou2025browsecomp}
Peilin Zhou, Bruce Leon, Xiang Ying, Can Zhang, Yifan Shao, Qichen Ye, Dading Chong, Zhiling Jin, Chenxuan Xie, Meng Cao, and 1 others. 2025.
\newblock Browsecomp-zh: Benchmarking web browsing ability of large language models in chinese.
\newblock \emph{arXiv preprint arXiv:2504.19314}.

\end{thebibliography}

\appendix
\onecolumn
\section{Appendix}

\subsection{Case Study: Demonstration of Our Deep Research Agent in Action}
\label{sec:agent-demonstration}
To demonstrate the effectiveness and trustworthiness of our Deliberative Search Agent, we present a case study based on a randomly sampled topic from the \textsc{DeepResearch} benchmark:
\textit{“What are the investment philosophies of Duan Yongping, Warren Buffett, and Charlie Munger?”}
This topic requires synthesis across multiple domains of knowledge, including individual investment principles, long-term financial strategies, and the historical context of major investment decisions.

\subsubsection{Planning and Decomposition}

Upon receiving the input topic, the \textbf{Planner Model} in our framework first decomposes the question into a structured set of \textit{Sections}, each targeting a key sub-area of the topic. This modular breakdown not only guides the research process but also ensures broad and deep coverage of the subject matter.

\begin{itemize}
\item \textbf{Section 1: Duan Yongping's Investment Philosophy} \\
\textit{Description:} An exploration of Duan Yongping's approach to investment, highlighting his strategic focus, key principles, and notable investments.

\item \textbf{Section 2: Warren Buffett's Investment Philosophy} \\
\textit{Description:} A detailed analysis of Warren Buffett's investment philosophy, including his value investing approach, famous quotes, and successful investment strategies.

\item \textbf{Section 3: Charlie Munger's Investment Philosophy} \\
\textit{Description:} A study on Charlie Munger's investment principles, with emphasis on his mental models, ideas about risk, and the concept of worldly wisdom.

\item \textbf{Section 4: Comparative Analysis of Philosophies} \\
\textit{Description:} A comparative analysis of the investment philosophies of Duan Yongping, Warren Buffett, and Charlie Munger, identifying common elements and key differences.

\end{itemize}

\subsubsection{Focused Query Generation}

Within each section, the \textbf{Researcher Module} further refines the information needs into targeted search queries. For instance, under the section “Duan Yongping’s Investment Philosophy,” the system automatically generated the following focused queries:

\begin{itemize}
\item \texttt{“Duan Yongping investment strategy principles”}
\item \texttt{“Notable investments by Duan Yongping and their outcomes”}
\end{itemize}

These queries enable the system to retrieve specific and relevant information rather than relying on generic content retrieval, ensuring higher factual grounding.

\subsubsection{Deliberative Search Output}

As an example, consider the query \texttt{“Duan Yongping investment strategy principles”}. Our \textbf{Deliberative Search Model} retrieved and synthesized relevant content through iterative reasoning and retrieval. Below is a representative output excerpt:

\begin{tcolorbox}[title = {Deliberative Search Model Output}, fontupper=\footnotesize, fontlower=\Large, breakable]
\begin{lstlisting}
<think>
Okay, so the user is asking about Duan Yongping's investment strategy principles. Let me start by recalling what I know. Duan Yongping is a Chinese entrepreneur and investor, co-founder of BBK Electronics, and known for his value investing approach. But I need to confirm the specifics of his strategy. Maybe he follows similar principles to Warren Buffett, since he's often compared to him. But I'm not sure. Maybe there are specific principles he's known for, like focusing on consumer goods, long-term investments, or certain financial metrics.
First, I should search for general information about his investment strategy. Let me try a search query.
</think>
<search>
Duan Yongping investment strategy principles
</search>
<result>
[{'id': 0, 'title': "Duan Yongping's Strategic Moves: A Closer Look at Apple Inc ...", 'abstract': 'Based in Menlo Park, California, Yongping is celebrated for his long-term investment philosophy, often focusing on technology and consumer ...'}, {'id': 1, 'title': "Duan Yongping's three major investment principles: do not ...", 'abstract': "Weekend reading | Duan Yongping's three major investment principles: do not short, do not borrow money, do not touch what you do not understand."}, {'id': 2, 'title': "Duan Yongping: don't do what you don't understand when ...", 'abstract': "... Duan Yongping's investment ideas. First, do not short, do not borrow money, do not do things you do not understand. To be honest, I don't ..."}, {'id': 3, 'title': "Duan Yongping's Business Ideas: Analysis of Three Core ...", 'abstract': 'Duan Yongping believes that excessive pursuit of "great goals" can easily lead to seeking quick success, adopting aggressive or even risky ...'}, {'id': 4, 'title': 'Pinduoduo plunges, Duan Yongping bottom fishing in a ...', 'abstract': "Duan Yongping's consistent investment principle is Value Investment. Value Investment favors companies and bottom fishing at low prices ..."}, {'id': 5, 'title': "The Great Way: Duan Yongping's Investment Q&A", 'abstract': '8. The risk of value investing should not be greater than that of driving out, and the risk of speculation is not smaller than going to a casino ...'}, {'id': 6, 'title': 'Doubling Down on Apple and Nvidia Amid AI Boom', 'abstract': "Duan's selective strategy in expanding his portfolio showcases a careful balance between responding to market volatility and capitalizing on ..."}, {'id': 7, 'title': "Duan Yongping's 20000-word transcript of his talk at ...", 'abstract': '- The essence of value investment: He believes that investment is essentially value investment, and that one should invest in companies that ...'}]
</result>
<CONFIDENCE:3>
<think>
Okay, so I got some search results. The titles and abstracts mention that Duan Yongping has three major principles: don't short, don't borrow money, and don't touch what you don't understand. Also, his approach is compared to value investing, similar to Warren Buffett. The third document's abstract specifically lists the three principles. The fourth document talks about his core ideas, and the seventh one is a transcript of a talk where he discusses value investing. Document 5 is a Q&A about his investment approach. Document 1 and 2 also mention the three principles. So it seems like the three principles are a key part of his strategy. But I need to confirm the details. Maybe reading document 1 or 2 would help. Document 1 is titled "Duan Yongping's three major investment principles..." so that's probably the best to read first.
</think>
<read>1</read>
<document 1>
Weekend reading | Duan Yongping's three major investment principles: do not short, do not borrow money, do not touch what you do not understand.
</document 1>
<CONFIDENCE:8>
<think>
Okay, so Document 1 clearly displays Duan Yongping's three major investment principles: do not short, do not borrow money, and do not touch what you do not understand. That's pretty straightforward. The title and the content both reinforce these three points. The other documents in the search results also mention these principles, so it seems consistent. Now, maybe I should check if there's any additional context or explanation in the other documents. For example, Document 2 has a similar title, and Document 5 is a Q&A which might provide more insights. But since Document 1 already lists the three principles explicitly, maybe that's sufficient. Alternatively, Document 7 is a 20,000-word transcript of a talk, which might have more detailed explanations. But since the user is asking for the principles, the three points from Document 1 are the key. I might not need to read more unless there's conflicting information. But from the abstracts, it seems the other documents support the same principles. So I can be confident that these three are his main principles. The rest of the information, like value investing, long-term focus, etc., are part of the context but the three rules are the core. So the answer should be those three principles.
</think>
<search>
Duan Yongping value investing approach
</search>
<result>
[{'id': 8, 'title': "Duan Yongping's Strategic Moves: A Closer Look at Apple Inc ...", 'abstract': 'Duan Yongping (Trades, Portfolio), a renowned investor known for his strategic and value-driven investment approach, recently submitted his ...'}, {'id': 9, 'title': "Duan Yongping: don't do what you don't understand when ...", 'abstract': 'My general goal is to buy what I think is undervalued by 50\% or more, and the value should be the discount of current net worth plus the sum of ...'}, {'id': 10, 'title': "Duan Yongping's three major investment principles: do not ...", 'abstract': "Weekend reading | Duan Yongping's three major investment principles: do not short, do not borrow money, do not touch what you do not understand."}, {'id': 11, 'title': "Duan Yongping's Strategic Moves: A Closer Look at Apple ...", 'abstract': 'Duan Yongping (Trades, Portfolio), a renowned investor known for his strategic and value-driven investment approach, recently submitted his ...'}, {'id': 12, 'title': 'Young Value Investor: How/what to read?', 'abstract': "You can find Duan's quote collections online, which I believe are essential reading for young Chinese value investors, worth pondering ..."}, {'id': 13, 'title': "Duan Yongping's Business Ideas: Analysis of Three Core ...", 'abstract': 'Duan Yongping has fully applied the concept of "do the right things, and do things right" in founding and investing in companies. Strategic Opportunity: In the ...'}, {'id': 14, 'title': "Duan Yongping's Options Trading Strategies", 'abstract': 'His options strategies are not simple speculation, but rather the application of derivatives based on the concept of value investing. 1. Core of ...'}, {'id': 15, 'title': 'Doubling Down on Apple and Nvidia Amid AI Boom', 'abstract': "Duan's selective strategy in expanding his portfolio showcases a careful balance between responding to market volatility and capitalizing on ..."}]
</result>
<CONFIDENCE:5>
<think>
Okay, so I searched again for Duan Yongping's value investing approach. The results include documents 8, 9, 10, etc. Document 8 and 11 mention his strategic and value-driven approach. Document 9 talks about buying undervalued assets by 50% or more. Document 10 is the same as the previous one. Document 13 discusses his core business ideas. Maybe Document 9 has more details on his value investing criteria. The abstract displays he aims to buy what's undervalued by 50% or more. That seems like a key part. So reading Document 9 might provide more context on how he applies value investing. Also, Document 14 talks about his options trading strategies based on value investing. But since the user is asking about the principles, maybe Document 9 is more relevant. Let me read Document 9.
</think>
<read>9</read>
<document 9>
My general goal is to buy what I think is undervalued by 50% or more, and the value should be the discount of current net worth plus the sum of ...
</document 9>
<CONFIDENCE:6>
<think>
Okay, so Document 9 displays that Duan Yongping's general goal is to buy what he thinks is undervalued by 50% or more. The value here is the discount of current net worth plus the sum of... Hmm, maybe the rest of the sentence is cut off, but from what's given, it seems he looks for significant discounts to intrinsic value. So combining that with the three principles from Document 1, his strategy emphasizes value investing with a margin of safety (50% undervalued), no shorting, no leverage, and sticking to what he understands. The other documents also support this. For example, Document 5 talks about the risk of value investing being comparable to driving, implying a conservative approach. Document 7's transcript might elaborate more, but since the key points are already clear, maybe that's enough. So the answer should be the three principles plus the emphasis on buying undervalued assets with a 50% margin of safety.
</think>
<search>
Duan Yongping investment examples
</search>
<result>
[{'id': 16, 'title': "Duan Yongping's Portfolio - H&H International Investment", 'abstract': "In Duan Yongping's portfolio as of 30 Jun 2025, the top 5 holdings are (AAPL) APPLE INC (62.47%), (BRK.B) BERKSHIRE HATHAWAY INC CL-B (14.24%), (PDD) PINDUODUO ..."}, {'id': 17, 'title': "Duan Yongping: don't do what you don't understand when ...", 'abstract': 'In 2001, Duan Yongping bought a large number of NetEase, Inc shares at a price of about US $1, making a maximum profit of a hundredfold.'}, {'id': 18, 'title': 'Duan Yongping holds these companies! The latest Share ...', 'abstract': 'According to the latest data submitted by H&H, as of the third quarter of 2024, it held 8 US stocks with a market value of 16.541 billion US dollars.'}, {'id': 19, 'title': "PDD's Chinese-American billionaire investor Duan Yongping ...", 'abstract': 'As of September 30, the US$16.54 billion portfolio of his H&H International Investment firm consisted of PDD, as well as Apple, Alphabet, ...'}, {'id': 20, 'title': 'Duan Yongping takes action! Splurges 2.9 billion to ...', 'abstract': 'Back to the five companies Duan Yongping is currently buying: NVIDIA, Apple, Taiwan Semiconductor, Google, and TENCENT.'}, {'id': 21, 'title': "Duan Yongping - Andy Lin's Long-term Stock Investment Blog", 'abstract': "Duan Yongping's investment empire and his BBK Group(Oppo, Vivo, One Plus, J&T Express, Realme) Almost no Taiwanese have heard of Duan Yongping, but it is not ..."}, {'id': 22, 'title': 'Duan Yongping holds these companies! The latest Share ...', 'abstract': 'Duan Yongping manages more than one stock account. He previously revealed on social media that his investments mainly include Apple in the US stock market, ...'}, {'id': 23, 'title': 'Value investment guru Duan Yongping: Building a US ...', 'abstract': "In this huge investment portfolio, technology giant Apple is undoubtedly Duan Yongping's favorite target, with a holding ratio of up to 80.95% ..."}]
</result>
<CONFIDENCE:4>
<think>
Okay, so I searched for Duan Yongping's investment examples. The results show that his portfolio as of June 30, 2025, had top holdings in Apple (62.47%), Berkshire Hathaway, Pinduoduo, etc. (Document 16). Document 17 displays he bought NetEase in 2001 at $1 and made a hundredfold profit. Document 19 displays H&H International Investment's portfolio included PDD, Apple, Alphabet. Document 20 lists companies he was buying in a splurge: NVIDIA, Apple, TSMC, Google, Tencent. Document 23 notes Apple as 80.95% of his portfolio. So his portfolio is heavily concentrated in tech and consumer companies, with Apple being the dominant holding. This aligns with his value investing approach, as Apple might be seen as undervalued relative to its fundamentals. Also, his investment in Pinduoduo (PDD) and Tencent (HKG:0700) fits the consumer sector focus. The example of NetEase shows his ability to identify undervalued companies early. But since the user is asking about principles, these examples support the principles by showing he sticks to what he understands (tech/consumer), avoids leverage (since his portfolio is cash-heavy or he doesn't borrow), and doesn't short. The concentration in a few stocks might also reflect his confidence in his analysis and the margin of safety he seeks.
</think>
<read>17</read>
<document 17>
In 2001, Duan Yongping bought a large number of NetEase, Inc shares at a price of about US $1, making a maximum profit of a hundredfold.
</document 17>
<CONFIDENCE:7>
<think>
Okay, so Document 17 displays that in 2001, Duan Yongping bought a large number of NetEase shares at around $1 and made a hundredfold profit. That's a significant example of his investing success. It reinforces the idea that he identifies undervalued companies (NetEase at $1) and holds them long-term for substantial gains. This example supports the principles of value investing with a margin of safety and sticking to what he understands. Also, his concentration in a few stocks like Apple and NetEase shows he's not diversified widely but focuses on a few bets he's confident in. No shorting or leverage here either. So all examples align with the previously identified principles.
</think>
<search>
Duan Yongping on diversification
</search>
<result>
[{'id': 24, 'title': "Duan Yongping's Business Ideas: Analysis of Three Core ...", 'abstract': 'Focus on Core Strengths, Reject Blind Diversification: No great ambition also means not being tempted by non-core "big opportunities" and ...'}, {'id': 25, 'title': "Duan Yongping's latest statement in 2025", 'abstract': 'Proper capital planning, diversified investments, and steady appreciation are the principles ordinary investors should follow. Editor/lambor.'}, {'id': 26, 'title': '(2/2) Duan the Dilettante', 'abstract': "Duan just announced his resignation from Subor. According to the agreement with Yihua, the parent company, Duan was entitled to 20% of Subor's profits."}, {'id': 27, 'title': "Duan Yongping - Andy Lin's Long-term Stock Investment Blog", 'abstract': "Duan Yongping's investment empire and his BBK Group(Oppo, Vivo, One Plus ... Diversified Conglomerate, AntiTrust, Business Diversity, Duan Yongping ..."}, {'id': 28, 'title': 'Value investment guru Duan Yongping: Building a US ...', 'abstract': 'Duan Yongping is known for his profound value investment philosophy. He is good at discovering high-quality assets that are undervalued by the market.'}, {'id': 29, 'title': 'Duan Yongping 2.90 billion bottom fishing god operation ...', 'abstract': 'This operation by Duan Yongping is by no means a simple bottom fishing, but a structured layout based on three core Value Investment principles. Volatility ...'}, {'id': 30, 'title': 'Doing the right thing is a lot less hassle! Duan Yongping shares ...', 'abstract': 'Diversification is generally due to the fact that the main business is not strong enough. They want to find a way through diversification, but as a result ...'}, {'id': 31, 'title': 'Duan Yongping continues to buy this company', 'abstract': 'Duan Yongping revealed that among his investments, US stocks are mainly in Apple, A-shares are mainly in Maotai, and Hong Kong stocks are mainly Tencent.'}]
</result>
<CONFIDENCE:3>
<think>
Okay, so I searched for Duan Yongping's views on diversification. The results are a mix. Document 24 displays he rejects blind diversification and focuses on core strengths. Document 25 displays his 2025 statement where he advises ordinary investors to follow proper capital planning, diversified investments, and steady appreciation. Document 30 depicts that diversification is generally because the main business isn't strong enough. Document 27 refers to BBK as a diversified conglomerate. Hmm, so there's a bit of a contradiction here. On one hand, he advises against blind diversification and focuses on core areas, but in another context, especially for ordinary investors, he suggests diversification. Maybe there's a distinction between his own concentrated portfolio (as seen in the examples) and his advice to others. Also, Document 31 shows his portfolio is mainly Apple, Maotai, Tencent---so concentrated. So his personal approach is concentrated, but he might advise others to diversify. Or maybe the context is different. Need to read Document 24 and Document 25 to clarify.
</think>
<read>24</read>
<document 24>
Focus on Core Strengths, Reject Blind Diversification: No great ambition also means not being tempted by non-core "big opportunities" and ...
</document 24>
<think>
Okay, Document 24 displays "Focus on Core Strengths, Reject Blind Diversification." So Duan Yongping advises against blind diversification, meaning not spreading investments too thinly without a clear reason. Instead, concentrate on areas where you have expertise or where the opportunity is clear. This aligns with his principle of not touching what you don't understand. His own portfolio is concentrated in a few stocks like Apple, Tencent, Maotai, which supports this. Document 25, which I haven't read yet, might have a different perspective, but based on Document 24, the emphasis is on avoiding unnecessary diversification. However, Document 25's abstract displays that proper capital planning, diversified investments, and steady appreciation are advised for ordinary investors. Hmm, so there's a possible nuance here. Maybe he distinguishes between blind diversification (bad) and strategic diversification (good). Or maybe Document 25 is more about general advice, while Document 24 is about his personal approach. I need to check Document 25 to see.
</think>
<read>25</read>
<document 25>
Proper capital planning, diversified investments, and steady appreciation are the principles ordinary investors should follow. Editor/lambor.
</document 25>
<think>
Okay, so Document 25 depicts that proper capital planning, diversified investments, and steady appreciation are principles ordinary investors should follow. This contrasts with Document 24's advice against blind diversification. So there's a distinction here. For ordinary investors, diversification is recommended as part of managing risk, while for Duan himself or those following his concentrated approach, sticking to core strengths and not diversifying blindly is key. The examples of his portfolio (high concentration) support that he practices a concentrated strategy. So his personal approach is concentrated, but he might advise others, especially ordinary investors, to diversify appropriately. This makes sense because individual investors might not have the same level of expertise or risk tolerance. So his investment principles for himself include concentration, while his advice for others includes diversification. But the question is about his own investment strategy principles, so the main points are the three core principles (no shorting, no leverage, stick to what you understand) and concentrated value investing with a significant margin of safety.
</think>
<answer>
Based on the available information, Duan Yongping's investment strategy principles [citation:1] emphasize three core tenets:  
1. Avoid Shorting: He advises against betting against companies through short-selling.  
2. No Leverage: Refraining from borrowing money to invest to mitigate risk.  
3. Understand What You Invest In: Focusing solely on areas within one's competence to reduce uncertainty.  
This approach is complemented by a value investing philosophy [citation:9], where he seeks companies undervalued by at least 50% relative to their intrinsic worth. His investment in NetEase in 2001 [citation:17] exemplifies this, yielding a hundredfold return. His portfolio as of 2025, dominated by Apple (62.47%) and a few other tech and consumer stocks [citation:16], reflects a concentrated strategy prioritizing deep understanding over diversification [citation:24].  
While he cautions against blind diversification [citation:24], he acknowledges that ordinary investors may benefit from prudent diversification [citation:25]. However, his personal approach prioritizes conviction in a select few high-quality assets.  
<CONFIDENCE:7>
</answer>
\end{lstlisting}
\end{tcolorbox}

From this example, we observe that the model follows a multi-round \textit{Think–Search–Read} cycle until a satisfactory answer can be produced. In each iteration, the model first reflects on its current information state (\textit{Think}), formulates refined search intents (\textit{Search}), and reads the retrieved content to update its knowledge base (\textit{Read}). Importantly, each round is accompanied by an internal estimation of confidence, followed by a calibration step to ensure the reliability of accumulated evidence before proceeding to the next stage.

\subsection{Case Study: Report Comparison}
\label{sec:report-Comparison}
For a randomly selected topic from the \textsc{DeepResearch} benchmark—\textit{“What are the investment philosophies of Duan Yongping, Warren Buffett, and Charlie Munger?”}—we compare the reports generated by different deep research agents, including \textbf{Perplexity} and \textbf{Doubao}.

\begin{tcolorbox}[title = {Ours Report}, fontupper=\footnotesize, fontlower=\Large, breakable]
\begin{lstlisting}
Investment Philosophies of Duan Yongping, Warren Buffett, and Charlie Munger: An Introduction
This report delves into the distinct yet interconnected investment philosophies of Duan Yongping, Warren Buffett, and Charlie Munger, three influential figures in the world of investing. Duan Yongping is noted for his commitment to understanding investments thoroughly and maintaining a conservative strategy, a principle seen in his significant investments in NetEase and Apple. Warren Buffett, known for investing in companies with enduring competitive advantages, exemplifies this with his investments in Coca-Cola and Apple. Meanwhile, Charlie Munger complements Buffett's strategy with a rational and multidisciplinary approach, emphasizing the importance of mental models. Together, these investors have shaped contemporary investment strategies, advocating for value investing and long-term success. 
The following sections will explore their philosophies in detail and illustrate their impact on modern investment practices.
## Overview of Duan Yongping's Investment Philosophy
Duan Yongping's investment philosophy is deeply rooted in the principles of value investing, drawing significant influence from Warren Buffett. His approach emphasizes several core tenets. Firstly, he strictly avoids investing in companies or sectors that he does not fully comprehend, underscoring the importance of clarity and understanding in investment decisions<CONFIDENCE:8>. Secondly, Duan's strategy is characterized by a conservative approach, eschewing shorting or leveraging to minimize risk<CONFIDENCE:8>. Thirdly, patience and a focus on long-term value are paramount. This is evidenced by his strategic investments in NetEase in 2001 and Apple in 2011, where he prioritized intrinsic value over short-term market fluctuations<CONFIDENCE:8>.
Duan's career clearly demonstrates the successful application of these principles. In 2001, he notably invested in NetEase at approximately $1 per share, leading to a substantial profit when the price soared to $70<CONFIDENCE:6>. His considerable investment in Apple, which comprised 63.33% of his portfolio as of March 2025, reflects his commitment to quality companies he comprehends, investing during times of undervaluation<CONFIDENCE:6>. Overall, Duan's investment strategy is marked by its simplicity, integrity, and a concentration in high-conviction holdings, aligning closely with Buffett's ideals of honest and transparent investing<CONFIDENCE:8>.
## Warren Buffett's Investment Philosophy
Warren Buffett's investment philosophy is founded on the principles of value investing, which emphasize acquiring undervalued companies with strong fundamentals and holding them for the long term. His strategy prioritizes companies with economic moats, such as sustainable competitive advantages and exemplary management quality<CONFIDENCE:5>. A prime example of this approach is his investment in Coca-Cola, initiated in 1988 with a $1.3 billion purchase. This reflected his belief in the brand's enduring value and competitive moat, focusing on high returns on capital and capital-light operations. As of 2025, Coca-Cola remained a significant holding, comprising 11.07% of Berkshire Hathaway's portfolio<CONFIDENCE:3>.
Buffett's investment in Apple in 2016 highlights his adaptability within his value investing framework. Initially an aversion to tech stocks, his $1 billion investment grew to represent 25.76% of the portfolio by leveraging Apple's share repurchases to enhance returns<CONFIDENCE:3>. This move demonstrates Buffett's recognition of Apple's strong consumer ecosystem, aligning with his emphasis on consumer-oriented businesses<CONFIDENCE:3>.
In 2023, Buffett exemplified his long-term strategy through a significant investment in Occidental Petroleum, acquiring 265 million shares valued at $11.7 billion. This stake rose to 27.8% by year-end, illustrating his focus on industries with predictable cash flows and resilience, driven by perceptions of undervaluation and potential in the energy sector<CONFIDENCE:5>. These examples underscore Buffett's strategy of identifying high-quality businesses, allocating substantial capital, and maintaining patience for sustainable growth<CONFIDENCE:3>.
## Charlie Munger's Investment Approach
Charlie Munger's investment philosophy is distinguished by his rational and multidisciplinary approach, which effectively complements Warren Buffett's strategies. Munger emphasizes investing in businesses within one's circle of competence that possess strong, durable competitive advantages (moats) and can be acquired at fair prices. This approach contrasts with Buffett's more expansive strategy but enhances it through Munger's cautionary perspective.<CONFIDENCE:4> Their partnership is rooted in shared core principles, such as valuing companies based on intrinsic worth and prioritizing long-term returns, but is enriched by Munger's skepticism and Buffett's optimism. This complementary dynamic supports Berkshire Hathaway's sustained success by pairing Buffett's exploratory vision with Munger's pragmatic discipline.<CONFIDENCE:4>
A significant component of Munger's philosophy is his emphasis on **mental models**, a method of decision-making that incorporates frameworks from various disciplines to address complex problems. Notable models include **Inversion**, considering problems in reverse to uncover potential failures;<CONFIDENCE:3> and the **Circle of Competence**, which involves focusing investments in areas of expertise to reduce uncertainty.<CONFIDENCE:3> Additionally, the **Lollapalooza Effect** highlights how factors can combine to produce significant results,<CONFIDENCE:3> while **Probability and Decision Trees** involve evaluating risks and potential outcomes through quantitative analysis.<CONFIDENCE:3> This integrated latticework of models empowers Munger to make informed, long-term investment decisions and challenges traditional investing norms by drawing from psychology, economics, and mathematics.<CONFIDENCE:3> Such a framework, though broad, is considered foundational to the successful value investing practiced by Munger and Buffett.<CONFIDENCE:3>



......

These insights underline their influence on modern investing, where prudent, disciplined strategies remain vital. For contemporary investors, assessing intrinsic value and embracing a long-term perspective, while also integrating newer analytical tools, aligns with their proven legacies.
\end{lstlisting}
\end{tcolorbox}

From this example, we illustrate the application of our \textbf{confidence assessment framework} for evaluating investment-related claims. The framework adopts a threshold-based model—assigning \textit{high confidence} to claims with a score above 6 and \textit{low confidence} to those below 4—primarily grounded in the verifiability and quality of supporting evidence.

High-confidence claims typically involve well-documented, widely accepted information. For instance, Duan Yongping's conservative investment strategy, exemplified by his 2025 portfolio allocation with 63.33\% in Apple, is supported by clear empirical data. Similarly, Warren Buffett's principle of economic moats, illustrated through canonical examples such as Coca-Cola, reflects long-standing, substantiated investment philosophy.

In contrast, low-confidence is attributed to claims that lack verifiable support or rely on subjective interpretation. Examples include the specific market conditions at the time of Duan’s entry into NetEase, the motivations behind Buffett’s increasing allocation to technology stocks, and the uncertain long-term trajectory of positions such as Occidental Petroleum in the volatile energy sector.

This framework explicitly links confidence levels to the availability of objective, high-quality evidence, enabling more calibrated reasoning and improving the reliability of the final research output.

\begin{tcolorbox}[title = {Perplexity's Report}, fontupper=\footnotesize, fontlower=\Large, breakable]
\begin{lstlisting}
Investment Philosophies of Duan Yongping, Warren Buffett, and Charlie Munger
Key Takeaway:
All three investors share a long-term, value-oriented approach rooted in rigorous analysis, integrity, and restraint---but each emphasizes distinct pillars: Duan Yongping stresses personal and corporate integrity and "doing what you understand", Warren Buffett focuses on intrinsic value, economic moats, and management quality, and Charlie Munger adds mental models, selective concentration, and the avoidance of common psychological traps.
 
Duan Yongping: Integrity and Focused Restraint
Duan Yongping's philosophy melds traditional value investing with an almost moral imperative:
- Integrity as Foundation
- Insists on "being a person of integrity" in every decision.
- Chooses investments in companies with honest cultures and trustworthy management--for example, Apple, Tencent, and Moutai.[1]
- Avoids unethical practices such as short-selling.
- "No Great Ambition": Pragmatic Steadiness
- Rejects blind expansion or scaling for its own sake.
- Values the health and longevity of a business over rapid growth.
- Core-Competence Focus
- "Do not touch what you do not understand"--only invests in industries where he has deep cognitive advantages.[2]
- Avoids both over-diversification and chasing hot but unfamiliar sectors.
- Win-Win Partnership Culture
- Delegates equity and autonomy to trusted partners, fostering loyalty and shared success.
- Believes "right business, right people," emphasizing entrepreneurs' integrity.[1]

Warren Buffett: Value, Moats, and Quality
Buffett's enduring principles, refined since his Graham days, center on buying great businesses at sensible prices:
- Intrinsic Value & Margin of Safety
- Seeks stocks trading below their intrinsic worth, providing a buffer against downside risk.[3]
- "Price is what you pay; value is what you get."
- Economic Moats
- Prioritizes companies with durable competitive advantages--brands, network effects, high barriers to entry.[4]
- Quality Management
- Invests in teams characterized by integrity, talent, and shareholder alignment.[4]
- Long Horizon & Patience
- Prefers multi-decade holds, famously quoting "our favorite holding period is forever."
- Views the market as a "voting machine" in the short term but a "weighing machine" in the long run.[3]
- Capital Preservation
- Rule No. 1: Never lose money; Rule No. 2: Never forget Rule 1.
- Sits on cash when no attractively valued opportunities exist.
- Simplicity & Circle of Competence
- Restricts investments to businesses he genuinely understands.
- Avoids complexity and market timing, focusing solely on fundamentals.

Charlie Munger: Mental Models, Selectivity, and Rationality
Munger's influence transformed Buffett's approach toward higher-quality businesses and sharper psychological awareness:
- Essential Filters
- Economic Moat: Prefers companies with sustainable competitive edges.
- Good Management: Integrity and talent are non-negotiable.
- Understandable Business Model: Simplicity aids correct valuation.
- Attractive Purchase Price: Fair price plus margin of safety.[5]
- Quality Over Bargains
- "Buy wonderful businesses at fair prices rather than fair businesses at wonderful prices."
- Concentration & Patience
- Maintains a focused portfolio (often fewer than ten holdings), betting big only on the best ideas.[6]
- "The big money is not in the buying and selling, but in the waiting."
- Psychological Discipline
- Emphasizes rationality and avoidance of emotional biases.
- Warns against both over- and under-diversification--advocates the sweet spot tailored to one's highest-conviction ideas.[7]
- Continuous Learning
- Leverages multidisciplinary mental models to sidestep common cognitive errors.
- Believes in learning from mistakes and market inefficiencies, often profiting when others err.

Conclusion:
While Duan Yongping, Buffett, and Munger all champion value investing and long-term orientation, they each highlight particular virtues--integrity and focus (Duan), economic moats and management quality (Buffett), and mental models, concentration, and psychological discipline (Munger)--that together form a comprehensive blueprint for disciplined, principled, and successful investing.
\end{lstlisting}
\end{tcolorbox}

\begin{tcolorbox}[title = {Doubao's Report}, fontupper=\footnotesize, fontlower=\Large, breakable]
\begin{lstlisting}
Duan Yongping, Warren Buffett, and Charlie Munger are three iconic investors with distinct yet interconnected philosophies. Here's a detailed analysis of their investment frameworks:
1. Warren Buffett: The Oracle of Omaha
Core Principles:
Buy businesses, not stocks: Buffett views stocks as fractional ownership of companies, focusing on long-term cash flow generation. His early "cigar butt" strategy (buying deeply undervalued companies) evolved into buying great businesses at fair prices, influenced by Charlie Munger .
Economic moats: He seeks companies with sustainable competitive advantages, such as brand loyalty (Coca-Cola), cost leadership (BNSF Railway), or network effects (Apple) .
Margin of safety: Buffett insists on purchasing companies at prices significantly below their intrinsic value, calculated via discounted cash flow (DCF) analysis .
Avoid leverage: He strongly opposes borrowing to invest, emphasizing that "leverage turns a 2-foot hurdle into a 10-foot wall" .
Key Practices:
Focus on the familiar: Buffett limits investments to industries he understands (e.g., consumer staples, insurance, utilities). His late embrace of Apple (2016) reflects a shift toward consumer tech with stable cash flows .
Long-term holding: Berkshire's top holdings (e.g., Coca-Cola, Apple) are held for decades, ignoring short-term market noise .
Capital allocation: Buffett prioritizes reinvesting profits into high-return opportunities or returning capital to shareholders via dividends and buybacks .
Criticisms/Adaptations:
His aversion to tech stocks for decades was seen as a limitation until Apple's success validated his revised approach.
Berkshire's massive size (>$1 trillion AUM) forces diversification, diluting his early "hyper-concentration" strategy .

2. Charlie Munger: The Philosopher of Rationality
Core Principles:
Latticework of mental models: Munger advocates integrating insights from physics, psychology, biology, and other disciplines to avoid narrow thinking. For example, he applies inversion ("What causes failure?") and critical mass theory (identifying tipping points in business growth) .
Avoid stupidity, not just seek brilliance: Munger focuses on eliminating errors (e.g., overpaying, ignoring management integrity) rather than chasing home runs .
Rationality and ethics: He emphasizes moral clarity, stating that "long-term greed" aligns with ethical behavior. This led Berkshire to avoid businesses like tobacco despite short-term profits .
Key Practices:
Focus on quality: Munger pushed Buffett toward buying great businesses at fair prices (e.g., Apple) over "cheap but mediocre" companies .
Psychological insights: He identifies cognitive biases (e.g., confirmation bias, herd mentality) to exploit market inefficiencies. For instance, he capitalized on panic during the 2008 financial crisis to invest in Goldman Sachs .
Multi-disciplinary analysis: When evaluating BYD (2008), Munger combined physics (battery technology), economics (China's growth), and psychology (trust in management) .
Legacy:
Munger's emphasis on intellectual rigor and ethical investing has deeply influenced Berkshire's culture, making it a model for long-term value creation .

3. Duan Yongping: The Chinese Value Investor
Core Principles:
Doing the right thing: Rooted in his Buddhist-inspired "ordinary mind," Duan prioritizes integrity and long-termism. He avoids speculative trends, adhering to no shorting, no borrowing, no investing in what you don't understand .
Business model and culture: He evaluates companies based on repeatable cash flow generation (e.g., Apple's ecosystem) and strong corporate cultures (e.g., Pinduoduo's customer-centric approach) .
Contrarianism: Duan capitalizes on market overreactions. His 2001 bet on NetEase (post-Internet bubble) at $0.25/share, later yielding 100x returns, exemplifies this .
Key Practices:
Extreme concentration: Over 80% of his portfolio is allocated to top 5 holdings (e.g., Apple, Berkshire, Alibaba), reflecting his confidence in high-conviction bets .
Industrial mindset: His experience founding BBK Electronics (OPPO, vivo) informs his focus on product quality and user loyalty. For example, he views Apple's iOS ecosystem as a "better moat than Coca-Cola's brand" .
Hybrid strategy: Duan combines Buffett's value principles with tech-sector agility. He uses options (e.g., selling puts on Apple) to enhance returns while maintaining long-term positions .
Challenges:
His reliance on a few sectors (tech, consumer) exposes him to regulatory risks, as seen in China's crackdown on internet companies .

Comparative Analysis
Aspect	Buffett	Munger	Duan Yongping
Core Focus	Moats and cash flow stability	Multidisciplinary rigor and ethics	Business model and culture
Risk Management	Avoid leverage; focus on intrinsic value	Inversion and checklist-driven decision-making	No leverage; "sell when wrong"
Sector Preference	Consumer staples, insurance, utilities	Diversified (e.g., BYD, Costco)	Tech, consumer, internet
Execution	Buy-and-hold; large-cap bias	Concentrated bets; hybrid of value and growth	Ultra-concentrated; small/mid-cap opportunities
Cultural Influence	American pragmatism	Western rationalism + Eastern wisdom	Chinese philosophy + Silicon Valley agility

Conclusion
While Buffett and Munger laid the groundwork for modern value investing, Duan Yongping has adapted their principles to tech-driven markets, emphasizing business model innovation and psychological resilience. All three share a commitment to long-termism, rationality, and avoiding irreversible mistakes. Their philosophies converge on the idea that investing is not about predicting markets but understanding what makes a business endure. As Duan succinctly puts it: "Investing is simple, but not easy--it requires patience and the courage to be different."
\end{lstlisting}
\end{tcolorbox}

\end{document}